\pdfoutput=1

\documentclass[11pt]{article}

\usepackage[preprint]{acl}

\usepackage{times}
\usepackage{latexsym}
\usepackage{amsmath}
\usepackage{amssymb}
\usepackage{pifont}
\usepackage{multirow}
\usepackage{booktabs}
\usepackage{tabularx}
\usepackage{xcolor}
\usepackage{colortbl}
\definecolor{mygrey}{rgb}{0.95,0.95,0.95}
\definecolor{myblue}{rgb}{0.9,0.9,1.0}
\definecolor{myyellow}{rgb}{1.0,0.9,0.85}

\newcommand{\dataset}{\textsc{HexaInst}\xspace}
\newcommand{\framework}{\textsc{SPARCOM}\xspace}

\usepackage[T1]{fontenc}

\usepackage[utf8]{inputenc}

\usepackage{microtype}

\usepackage{inconsolata}

\usepackage{graphicx}
\usepackage{amsmath}
\usepackage{subcaption}
\usepackage{microtype}
\usepackage{inconsolata}
\usepackage{graphicx}
\usepackage{amsmath}
\usepackage{amssymb}
\usepackage{algorithm}
\usepackage{algorithmic}
\usepackage[listings,skins,breakable]{tcolorbox}
\usepackage{float}
\usepackage{xspace}  
\usepackage{enumitem} 
\usepackage{pifont}

\title{Unveiling Instruction-Specific Neurons \& Experts: \\An Analytical Framework for LLM's Instruction-Following Capabilities}

\author{Junyan Zhang\textsuperscript{\rm 1,*}, Yubo Gao\textsuperscript{\rm 1,*}, Yibo Yan\textsuperscript{\rm 1,\rm 2}, Jungang Li\textsuperscript{\rm 1}, \textbf{Zhaorui Hou}\textsuperscript{\rm 1},\\
\textbf{Sicheng Tao}\textsuperscript{\rm 1},
\textbf{Shuliang Liu}\textsuperscript{\rm 1,\rm 2},
\textbf{Song Dai}\textsuperscript{\rm 1},
\textbf{Yonghua Hei}\textsuperscript{\rm 1},
\textbf{Junzhuo Li}\textsuperscript{\rm 1,\rm 2},
\textbf{Xuming Hu}\textsuperscript{\rm 1,\rm 2,}\footnotemark[2]\\
\\
\textsuperscript{\rm 1}The Hong Kong University of Science and Technology (Guangzhou)\\
\textsuperscript{\rm 2}The Hong Kong University of Science and Technology \\
\texttt{\{junyanzhang0317, yubogao1015\}@gmail.com}, \texttt{xuminghu@hkust-gz.edu.cn}
}

\begin{document}
\maketitle
\renewcommand{\thefootnote}{\fnsymbol{footnote}}
\footnotetext[1]{Equal contribution.}
\footnotetext[2]{Corresponding author.}
\renewcommand{\thefootnote}{\arabic{footnote}}
\begin{abstract}
The finetuning of Large Language Models (LLMs) has significantly advanced their \textit{instruction-following capabilities}, yet the \textit{underlying computational mechanisms driving these improvements remain poorly understood}. This study systematically examines how fine-tuning reconfigures LLM computations by isolating and analyzing instruction-specific sparse components, \textit{i.e.,} neurons in dense models and both neurons and experts in Mixture-of-Experts (MoE) architectures. In particular, we introduce \textbf{\dataset}, a carefully curated and balanced instructional dataset spanning six distinct categories, and propose \textbf{\framework}, a novel analytical framework comprising three key contributions: \ding{182} a method for identifying these sparse components, \ding{183} an evaluation of their functional generality and uniqueness, and \ding{184} a systematic comparison of their alterations. Through experiments, we demonstrate functional generality, uniqueness, and the critical role of these components in instruction execution. By elucidating the relationship between fine-tuning-induced adaptations and sparse computational substrates, this work provides deeper insights into how LLMs internalize instruction-following behavior for the trustworthy LLM community.

\end{abstract}

\section{Introduction}

Large Language Models (LLMs) fine-tuning has significantly enhanced the ability of LLMs to comprehend user intent, follow instructions, and align with human preferences, thereby boosting their performance across diverse tasks \citep{prakash2024fine,dang2024exploring,yan2025position,zhang2023instruction}. However, precisely \textit{how these fine-tuning processes alter the internal computational mechanisms of models to achieve superior instruction following remains a crucial yet elusive scientific question}. To shed light on these mechanisms, this study adopts an interpretability perspective. 

\begin{figure}[ht]
    \centering 
    \includegraphics[width=0.5\textwidth]{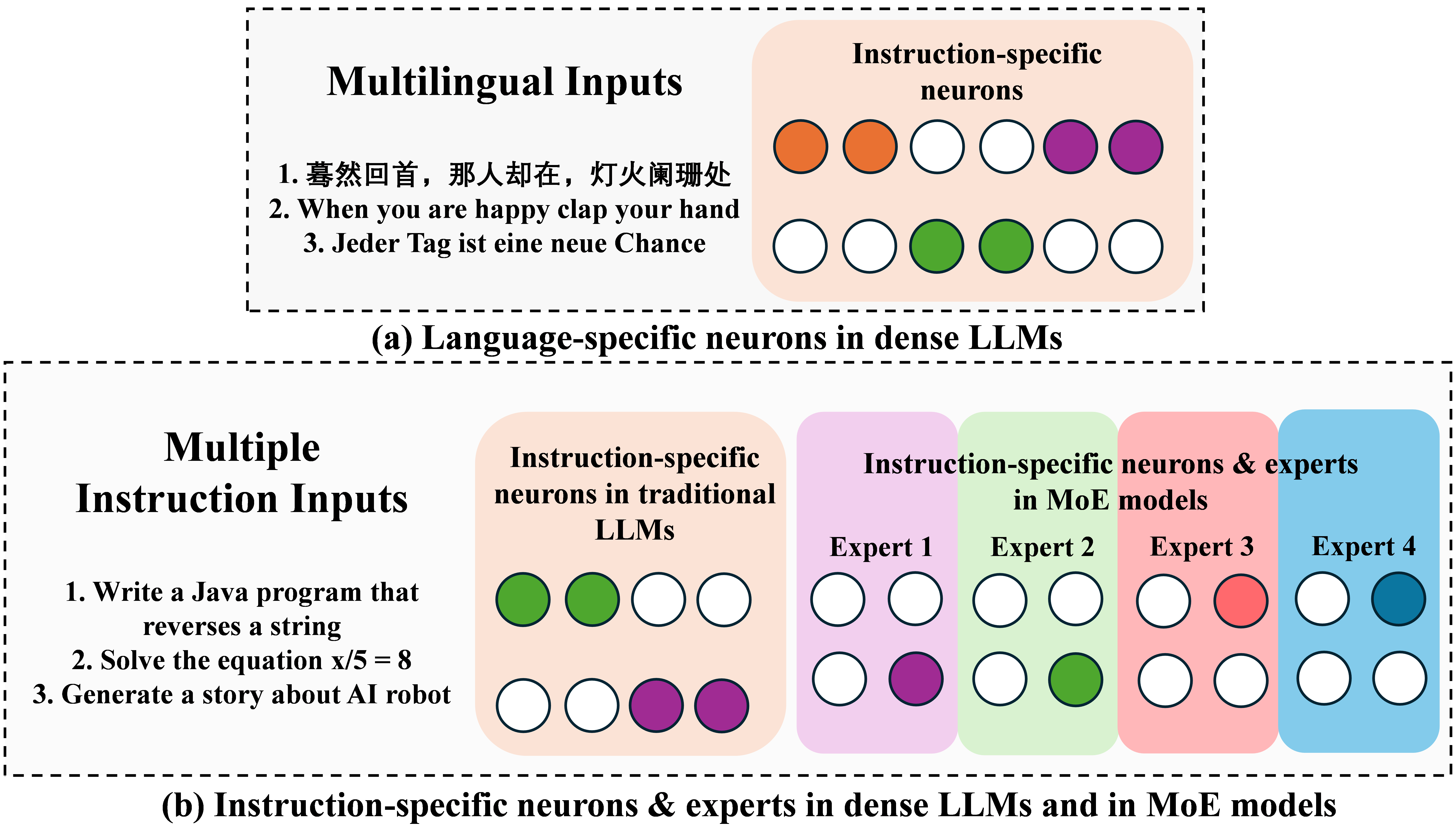} 
    \caption{Comparison of research focuses between Language-Specific Neurons (a) and Instruction-Specific Neurons \& Experts in dense LLMs \& MoE models (b).}
    \label{fig:intro} 
    \vspace{-4mm}
\end{figure}

Prior works in neuron-level interpretation \citep{tang2024language, huo2024mmneuron, kojima2024multilingual, wang2022finding, huang2024miner} has successfully identified \textit{X-specific neurons} crucial for storing factual knowledge \citep{dai2021knowledge}, processing specific languages \citep{tang2024language}, recalling domain information \citep{huo2024mmneuron}, and ensuring model safety \citep{chen2024finding}. These specific neurons make up a small proportion, but they are crucial to the model's corresponding capabilities, as illustrated in Figure \ref{fig:intro} (a). In the field of circuit analysis, it has been demonstrated that there exist several important circuits that store specific knowledge for particular tasks, such as indirect object identification \citep{wang2022interpretability} and color object identification \cite{merullo2023circuit}. Inspired by these findings, we posit a central hypothesis: \textbf{\textit{Does the remarkable instruction-following capability exhibited by instruction-tuned models also stem from certain sparse components? }}

In this work, we conduct a systematic investigation into the activation mode of two types of sparse components: \textbf{Instruction-Specific Neurons} and \textbf{Instruction-Specific Experts} within two popular open-source LLM families (LLaMA and Mistral) and one Mixture-of-Experts (MoE) model family (Qwen-MoE), as shown in Figure \ref{fig:intro} (b). Therefore, we first propose a meticulously curated and balanced instructional \textbf{\dataset} dataset comprising six instruction categories. We further propose \textbf{\framework}, a novel \textbf{\underline{spar}}se \textbf{\underline{com}}ponent analysis framework with three key steps. \ding{182} We identify Instruction-Specific Neurons and Instruction-Specific Experts within LLMs, enabling precise localization of these sparse components. These neurons and experts are primarily responsible for processing and executing instructions. \ding{183} We evaluate the generality and uniqueness of the distribution of the identified Instruction-Specific Neurons and Experts, providing a robust methodology for assessing their functional characteristics. \ding{184} We perform an alteration comparison, analyzing the differences of Instruction-Specific Neurons and Experts in the same model before and after fine-tuning. Also, we examine the distribution patterns of Instruction-Specific Neurons across different layers and propose a three-stage framework for understanding the internal mechanism. Ultimately, we have obtained several significant findings, offering new insights into how fine-tuning shapes the internal mechanisms of LLMs.

Contributions can be summarized as follows\footnote{Code and dataset will be released upon acceptance.}:
\begin{itemize}[leftmargin=*]
    \item We propose a meticulously curated and balanced \textbf{\dataset} dataset comprising six instruction categories for our in-depth analysis. 
    \item We present \textbf{\framework}, a novel framework designed to identify and analyze instruction-specific neurons and experts in LLMs. 
    \item We explore the generality and uniqueness of these specialized sparse components and uncover how fine-tuning shapes LLMs through them, revealing their distribution and activation patterns.
\end{itemize}

\section{Preliminaries and Related Works}

\subsection{Preliminaries}

\paragraph{Dense LLMs}
The conventional decoder-only transformer model takes an input sequence of tokens $ t = (t_1, t_2, \ldots, t_n) $, where $ t \in V^n $, and transforms it into an output probability distribution $ y = (y_1, y_2, \ldots, y_n) $, with $ y \in \mathbb{R}^{n \times |V|} $. Let $ x_i^{(l)}(t) \in \mathbb{R}^{d_{\text{model}}} $ denote the residual stream activation for the token at position $ i $ at the start of the $ l $-th layer. The transformation at each layer includes two main components: Attention Mechanism and Feed-Forward Network (FFN):

\begin{equation}
    \tilde{x}_i^{(l)} = x_i^{(l)} + \text{Attn}^{(l)}(x_{1:i}^{(l)}).
\end{equation}

\begin{equation}
    x_i^{(l+1)} = \tilde{x}_i^{(l)} + \text{FFN}^{(l)}(\tilde{x}_i^{(l)}).
\end{equation}

In FFN component, we do the following:

\begin{equation}
\begin{aligned}
y &= W_{\text{down}} \cdot \Bigg( 
    \text{act\_fn}\bigg(W_{\text{gate\_up\_proj}} \cdot \tilde{x}_i^{(l)}[0:d_{\text{mid}}] \bigg) \\
&\quad \odot \bigg(W_{\text{gate\_up\_proj}} \cdot \tilde{x}_i^{(l)} \bigg) [d_{\text{mid}}:2d_{\text{mid}}]
\Bigg),
\end{aligned}
\end{equation}

where \( W_{\text{gate\_up\_proj}} \) and \( W_{\text{down}} \) are learnable weight matrices. The \texttt{gate\_up\_proj} is used to project the input into a higher-dimensional space, dividing it into a gating part and an up-projection part. Then, the gating part passes through an activation function to determine the flow of information. To streamline the representation, the bias terms have been omitted from the formulation.

\begin{figure*}[ht]
    \centering 
    \includegraphics[width=0.8\textwidth]{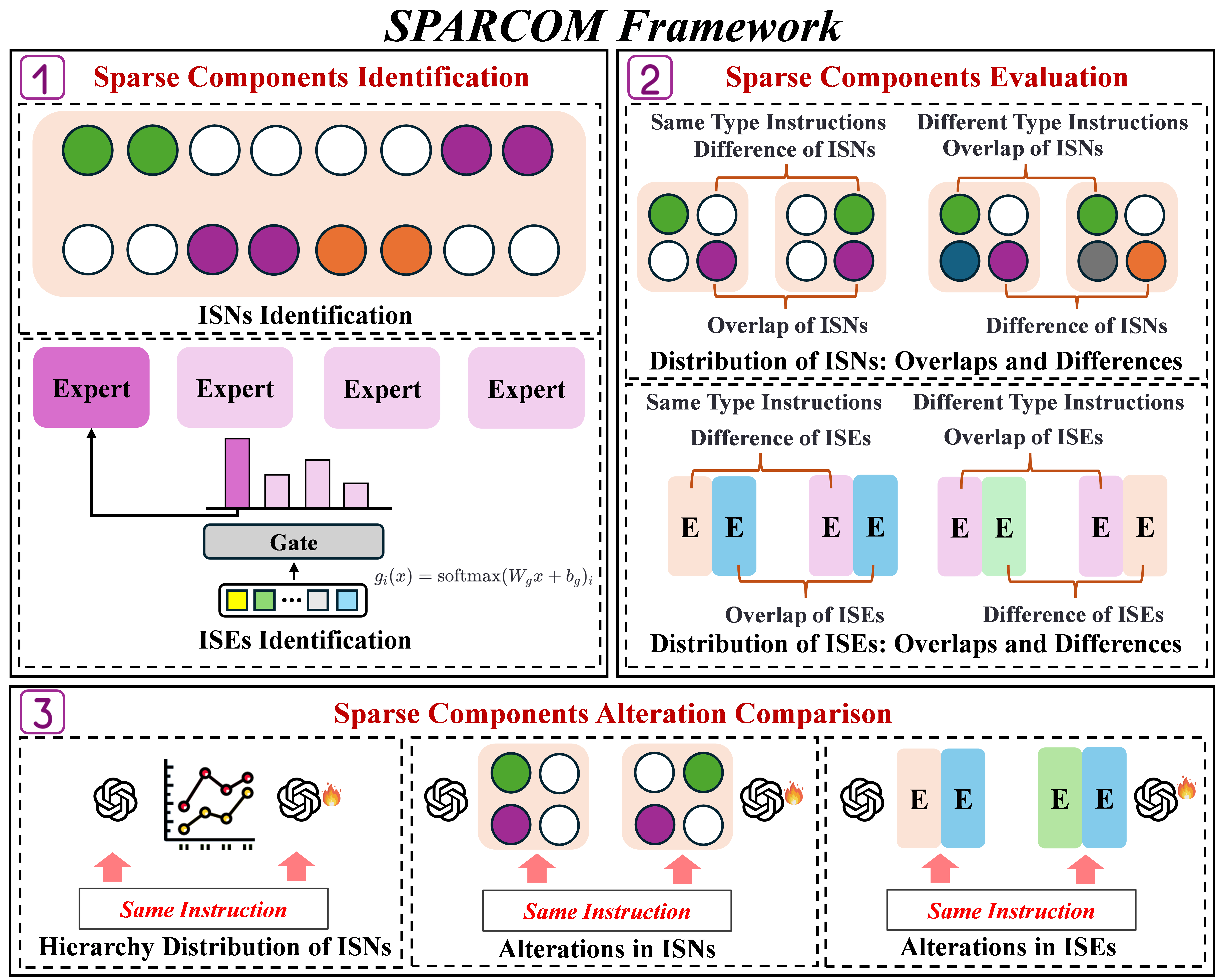}
    \caption{The \framework framework, which comprises three elements, aims for the identification \& evaluation of sparse components. ISNs and ISEs denote Instruction-Specific Neurons and Instruction-Specific Experts.}
    \label{fig:trend} 
    \vspace{-4mm}
\end{figure*}

\paragraph{MoE models}

In MoE models, the FFN is redesigned with a gating network to select multiple experts per token, incorporating a load balancing mechanism to optimize performance and computational load while ensuring all experts contribute fairly. Specifically, the FFN layer is replaced with the following formula:

\begin{equation}
y = \sum_{j \in \text{Top-k}(g)} g_j \cdot \text{FFN}_j(\tilde{x}_i^{(l)}),
\end{equation}

where $ g $ is the weight vector output by the gating network. In our implementation, we use the Qwen-MoE models. 

\subsection{Related Work}

\paragraph{MoE Models}

The MoE approach employs individual, independent experts to handle different tasks \cite{jacobs1991adaptive, jordan1994hierarchical}. In recent years, with the advancement of LLMs, MoE has emerged as an effective method for significantly scaling up model capacity while minimizing computational overhead, thereby attracting increasing attention from both academia and industry \cite{huang2025ultrasparsememorynetwork,cai2025survey,lin2024moe}. The core idea behind MoE is to introduce conditional computation: instead of applying the same parameters to all inputs, different inputs are processed by different parts of the model. This allows for scalable and efficient model growth \cite{shi2024time,li2025uni,yuan2025moe, yang2025faster}.

\paragraph{LLM Fine-tuning}

While foundational pre-trained language models possess extensive knowledge and certain reasoning capabilities \cite{ke2025survey,yan2024errorradar,yan2024survey}, they often struggle to directly meet diverse and specific user needs \cite{kumar2025llm,wei2025survey,li2025system}. To bridge this gap, researchers widely explore fine-tuning techniques to adapt models to general domains \cite{han2024parameter,wang2024datasurvey,wang2024parameter}. This typically involves two key components. Supervised fine-tuning enables the model to learn and generalize across a wide range of general instructions \citep{wei2022finetunedlanguagemodelszeroshot, wang2023selfinstructaligninglanguagemodels,shen2024tag}, while alignment with human preferences ensures that the model's behavior is refined to better adhere to human values and expectations \citep{christiano2023deepreinforcementlearninghuman, stiennon2022learningsummarizehumanfeedback, rafailov2024directpreferenceoptimizationlanguage,wang2024comprehensive,ji2024aligner}.

\paragraph{Neuron Analysis}
In recent years, mechanistic interpretability has gained prominence \cite{marks2024sparse, yao2024knowledge,cambria2024xai,bilal2025llms,mumuni2025explainable,wu2024usable}, with neuron analysis emerging as a powerful approach for uncovering the internal mechanisms of LLMs \cite{wang2024sharing, song2024does}. Recent studies in this area have successfully identified neurons that are either language-specific or domain-specific, thereby uncovering the specialized roles that certain neurons play within these models \cite{tang2024language, huo2024mmneuron, chen2024finding}. In our work, we apply neuron analysis techniques to specific instructions, with a particular focus on understanding how activated neurons before and after instruction tuning.

\section{\framework Framework}

\textbf{\framework} introduces an innovative sparse components analysis framework composed of three core steps, as depicted in Figure \ref{fig:trend}. The initial step focuses on pinpointing Instruction-Specific Neurons (ISNs) and Instruction-Specific Experts (ISEs) in LLMs, allowing for accurate detection of these sparse, task-aligned components. Leveraging this identification, the second step systematically analyzes the generality and uniqueness of the discovered neurons and experts, establishing a rigorous approach to evaluating their functional properties. In the third step, we analyze how the activation distributions of these sparse components change before and after model fine-tuning.

\subsection{Sparse Components Identification}

\paragraph{ISEs Identification}
\label{sec:Instruction Specific Neurons}
Inspired by the language activation probability entropy proposed by \citet{tang2024language}, we developed a method to identify ISNs.

Specifically, for each instruction $I$, we perform the following procedure. The activation of neurons mainly focuses on the Feed-Forward Network component. We feed $I$ into the LLM and record the activation value of each neuron $i$ in $j$-th \texttt{gate\_up\_proj} layer after applying the activation function:

\begin{equation}
    A_{ij}^{t}(I) = f\left(\text{gate\_up\_proj}(\tilde{x}_t^{(j)}(I)\right)_{i},
\end{equation}
where $f(\cdot)$ denotes the activation function applied to the layer's output of the token \( t \) across the instruction sequence length \( T \).

The activation frequency is empirically estimated by the likelihood that the neuron’s activation value exceeds zero:

\begin{equation}
    p_{ij}(I) = \frac{1}{T} \sum_{t=1}^{T} [A_{ij}^t(I) > 0].
\end{equation}

Subsequently, the activation frequency of each neuron in response to a given instruction is calculated. This process involves flattening all neurons across every layer and computing their respective activation frequencies. A threshold is then established to identify the top $\epsilon$ percentile of these frequencies. Neurons exceeding this threshold are designated as ISNs, reflecting their heightened propensity for activation in response to the specific instruction:

\begin{equation}
S(I) = \{(i, j) \mid p_{ij} \geq \epsilon \}
\end{equation}

For MoE models, we use similar ways to calculate the activation frequency:

\begin{equation}
    p_{eij}(I) = \frac{1}{T} \sum_{t=1}^{T} \left[A_{eij}^t(I) > 0\right] \cdot \left[e \in E_{jt}\right],
\end{equation}

where $e$ represents the index across all experts and $E_{jt}$ represents the top-k activated experts of token t in $j$-th layer.

The ISNs of MoE models can be calculated as the following:

\begin{equation}
S(I) = \{(e, i, j) \mid p_{eij} \geq \epsilon\}.
\end{equation}

\paragraph{ISEs Identification}

In MoE models, for each token, the routing mechanism selects the top-k experts from the output probability distribution, thereby determining the chosen experts. We name the activated experts as ISEs.

\subsection{Sparse Components Evaluation}

\paragraph{Distribution of ISNs: Overlaps and Differences}

We have already identified the ISNs, and we want to explore whether there is overlap in the distribution of these ISNs among instructions of the same type, and whether there are significant differences between different types of instructions:

\begin{equation}
\begin{aligned}
Sim(m_1, m_2) = & \frac{1}{|N_{m_1}||N_{m_2}|} \cdot \\
&\quad \sum_{\substack{n_1 \in m_1 \\ n_2 \in m_2}} J(S(I_{m_1n_1}), S(I_{m_2n_2})),
\end{aligned}
\end{equation}

where, \( \text{Sim}(m_1, m_2) \) represents the similarity of ISNs between two types of instructions, \( m_1 \) and \( m_2 \) represent the types of instructions, and \( n_1 \) and \( n_2 \) represent the indices of the specific instructions within their respective types, $J$ represents jaccard similarity. For MoE models, we use the similar methods to calculate.

This metric reflects the overlaps and differences in the distribution of ISNs and between same-type and different-type instructions. 

\paragraph{Distribution of ISEs: Overlaps and Differences}

For instruction \( I \), we collect the activation frequencies of all experts across each layers:
\begin{equation}
f_{ej}(I) = \frac{1}{T} \sum_{t=1}^{T} x_{ejt}, \quad f_{ej}(I) \in [0, 1]
\end{equation}

where $x_{ejt}$ represents whether the $t$-th token in the $j$-th layer activates the $e$-th expert.

After this, we flatten the obtained activation probability matrix into a one-dimensional vector:
\begin{equation}
F(I) = [f_{ej}(I)],
\end{equation}

Given two instructions \( I_1 \) and \( I_2 \), along with their activation frequency vectors \( F(I_1) \) and \( F(I_2) \), calculate the pearson correlation coefficient between them: 

\begin{equation}
\begin{split}
r_{I_1I_2} = {} 
& \frac{\sum_{i} \left[F(I_1)[i] - \overline{F(I_1)}\right] \left[F(I_2)[i] - \overline{F(I_2)}\right]}{\sqrt{\sum_{i} \left[F(I_1)[i] - \overline{F(I_1)}\right]^2}} \\
& \cdot \frac{1}{\sqrt{\sum_{i} \left[F(I_2)[i] - \overline{F(I_2)}\right]^2}}.
\end{split}
\end{equation}

Next, compute the average pearson correlation coefficient for all pairs of instructions from these two different or identical types:

\begin{equation}
Corr_{m1m2} = \frac{1}{|N_{m_1}||N_{m_2}|} \sum_{n_{1}=1}^{N_{m1}} \sum_{n_{2}=1}^{N_{m2}} r_{I_{m_{1}n_{1}}I_{m_{2}n_{2}}}.
\end{equation}

This metric reflects the correlation in the distribution of activated experts between same-type and different-type instructions.

\subsection{Sparse Components Alteration Comparison}

In this chapter, we evaluate alterations in ISNs and ISEs from three perspectives.

\paragraph{Hierarchy Distribution of ISNs}

We visualize the ISNs across layers of the model before and after fine-tuning, based on the computed results.

\paragraph{Alterations in ISNs}

We compare the jaccard similarity of distribution changes of the same instruction of activated neurons in the model before and after fine-tuning. * denotes that the calculation is between the model before and after fine-tuning:
\begin{equation}
    Sim(m) = \frac{1}{|N|} \sum_{n=1}^{|N|} J^*\left(S(I_{mn}), S(I_{mn})\right).
\end{equation}

\paragraph{Alterations in ISEs}

Similarly, we compute the average pearson correlation coefficient for all pairs of the same instruction in the model before and after fine-tuning:

\begin{equation}
    {Corr}_{m} = \frac{1}{|N|} \sum_{n=1}^{|N|} r^*_{I_{mn} I_{mn}}.
\end{equation}

\section{Experiments}

\subsection{Experimental Setups}

\paragraph{Models}
We conducted our study primarily on three families of publicly available LLMs: LLaMA \citep{touvron2023llama}, Mistral \citep{Jiang2023Mistral7}, and Qwen \citep{bai2023qwen}. For fine-tuned models, we examined multiple versions of LLaMA-2-Chat, specifically the 7B and 13B variants, along with Mistral-7B-Instruct-v0.1 and Qwen1.5-MoE-A2.7B-Chat. For the vanilla models, we selected LLaMA-2-7B, LLaMA-2-13B, Mistral-7B-v0.1, and Qwen1.5-MoE-A2.7B.

\paragraph{Datasets} 

Our work requires a dataset containing various types of instructions, which must be clear and precise, and preferably free of any extraneous information that could cause interference. However, current instruction datasets are \textit{unevenly distributed across tasks}, particularly lacking summarization and classification instructions, as well as AI-generated ones. We construct a balanced dataset \textbf{\dataset} with 1,200 instances across six instruction categories: \textit{classification} (CLS), \textit{code} (CODE), \textit{generalqa} (QA), \textit{generation} (GEN), \textit{math} (MATH), \textit{summarization} (SUM). Each category contains 100 AI-generated and 100 human-curated instructions to control for source variability. Specifically, the dataset is compiled through two primary sources. 
\textbf{Synthetic data:} Generated via \texttt{DeepSeek R1} with constrained meta-prompts. \textbf{Natural data:} Built upon public benchmarks:

\begin{itemize}[leftmargin=*]
    \item Classification: FLAN Collection \cite{longpre2023flan} 
    \item Code: HumanEval \cite{chen2021evaluating}
    \item GeneralQA: TriviaQA \cite{2017arXivtriviaqa}
    \item Generation: Alpaca \cite{alpaca}    
    \item Math: Math-500 \cite{lightman2023lets}
    \item Summarization: FLAN Collection \cite{longpre2023flan} 
\end{itemize}

For natural data, we extract instructions using regex pattern matching followed by expert validation and refinement. Synthetic instructions are cross-checked against training data of public LLMs to prevent contamination. The balanced design (100 synthetic plus 100 natural per category) enables disentangling neuron activation patterns from data source biases. All data have undergone manual post-validation to ensure quality. The details of post-validation can be found in \S\ref{appendix: validation}. Examples of each type of instructions can be found in \S\ref{appendix: datasets}.

\paragraph{Implementation Details}

We use the vllm \cite{kwon2023efficient} and Transformer library to obtain and hook the internal states of LLMs. For Qwen1.5-MoE-A2.7B-Chat and Qwen1.5-MoE-A2.7B model, we use the default settings, which involve selecting four dynamic experts for each token from a pool of sixty experts based on the gating network’s scores. 

\begin{figure}[t]
    \centering 
    \includegraphics[width=0.48\textwidth]{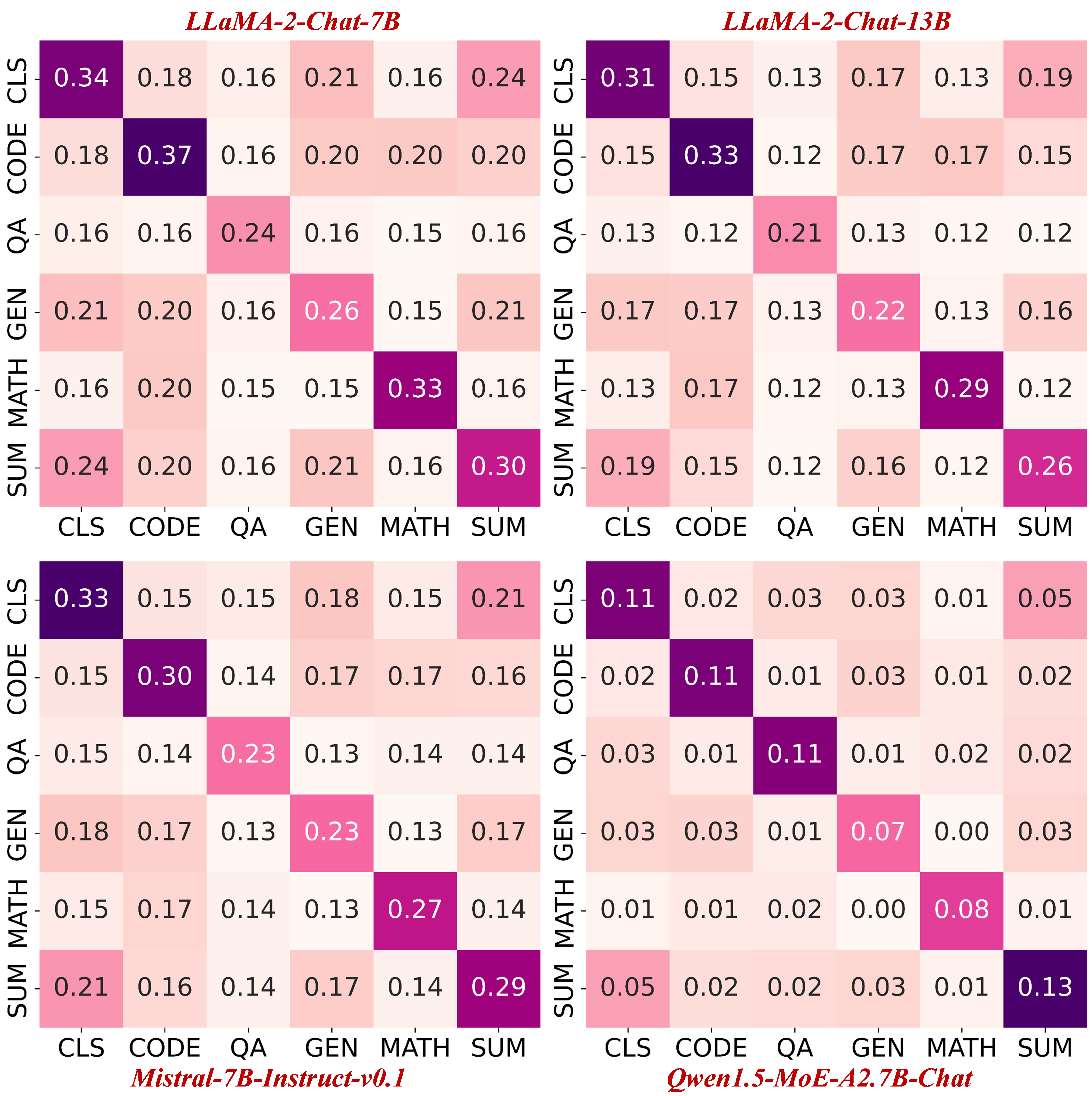} 
    \caption{Overlaps and differences in ISNs distribution across same-type and different-type instructions on LLaMA-2-Chat-7B, LLaMA-2-Chat-13B, Mistral-7B-Instruct-v0.1, and Qwen1.5-MoE-A2.7B-Chat.}
    \label{fig:jaccard}
    \vspace{-4mm}
\end{figure}

\begin{figure}[t]
    \centering 
    \includegraphics[width=0.35\textwidth]{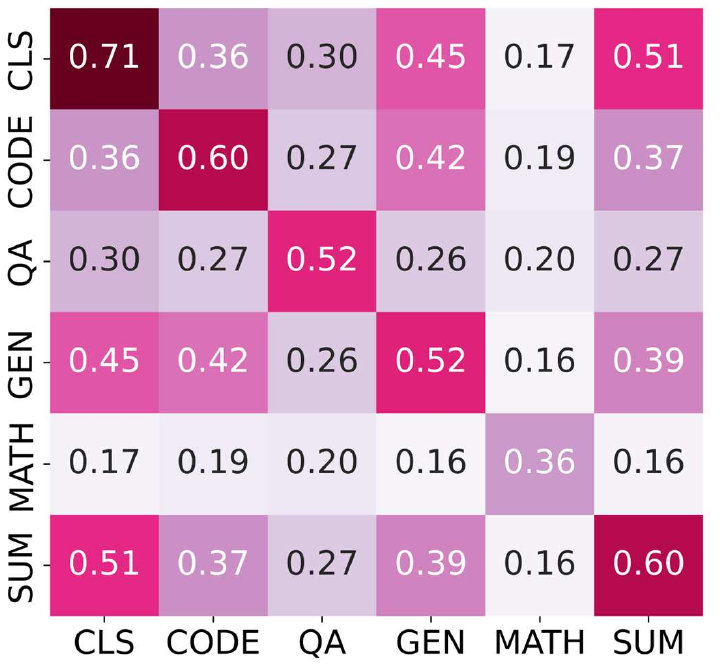}
    \caption{Overlaps and differences in ISEs distribution across same-type and different-type instructions on Qwen1.5-MoE-A2.7B-Chat.}
    \label{fig:expert} 
    \vspace{-4mm}
\end{figure}

\section{Results and Insights}

Through experiments, we conducted three key steps of our framework: \textbf{Sparse Components Identification}, \textbf{Sparse Components Evaluation}, and \textbf{Sparse Components Alteration Comparison}. Based on the results, we derived meaningful insights and significant findings regarding the behavior and characteristics of instruction-specific components in LLMs.

\subsection{Generality and Uniqueness of Sparse Components}

We propose that the ISNs and ISEs identified through \textbf{\framework} framework can be categorized into two types: general and unique ISNs, and general and unique ISEs.

\begin{figure*}[t]
    \centering
    \resizebox{0.9\textwidth}{!}{\includegraphics{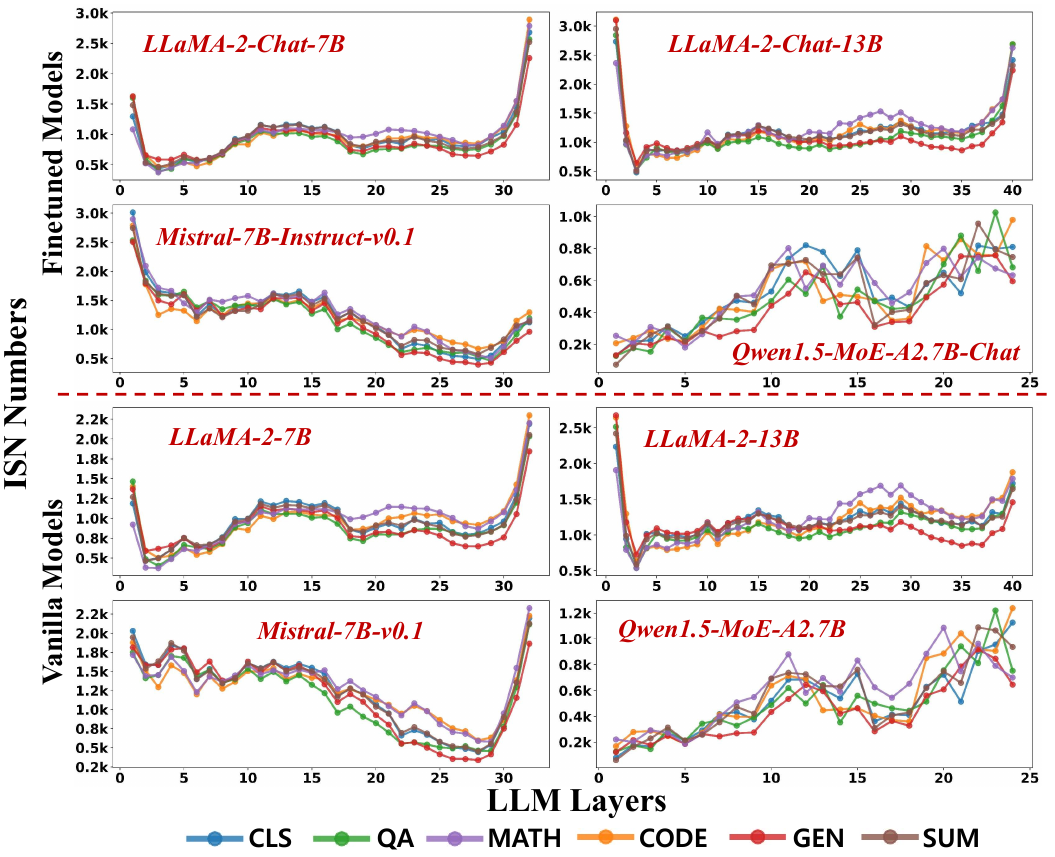}}
    \caption{Hierarchy distribution of ISNs across different layers. The upper part includes LLaMA-2-Chat-7B, LLaMA-2-Chat-13B, Mistral-7B-Instruct-v0.1, and Qwen1.5-MoE-A2.7B-Chat models. The down part includes LLaMA-2-7B, LLaMA-2-13B, Mistral-7B-v0.1, and Qwen1.5-MoE-A2.7B models.}
    \label{fig:ISN_distribution}
    \vspace{-4mm}
\end{figure*}

\paragraph{Generality}

As shown in Figure \ref{fig:jaccard}, we believe that the overlapping neurons between different instruction types mainly belong to general ISNs. These neurons are responsible for processing general instruction language and encode the common functions or conceptual elements required for instruction processing, or handle parts unrelated to the specific content of the instructions. For example, although there are clear semantic and expressive differences between different types of instructions, overlap in certain vocabulary is inevitable. This also leads to the overlap of general neurons across different instruction types. The high overlap of ISNs between \textit{classification} and \textit{summarization} likely reflects their intrinsic connection and shared skill requirements.

Similarly, in Figure \ref{fig:expert}, there is also a correlation in the activation of experts between different instruction types, although it is much weaker compared to the correlation within the same type of instructions. We believe that these general experts may be responsible for handling general instructions and responding to potential overlaps in tokens across different instructions.

\paragraph{Uniqueness}

As illustrated in Figure \ref{fig:jaccard}, all models exhibit a notably darker coloration along the diagonal, particularly for instruction types such as \textit{classification}, \textit{summarization}, \textit{code}, and \textit{math}. This indicates a high overlap of ISNs among instructions of the same type. Despite considerable variations in vocabulary and syntax within the same category of instructions, a pronounced similarity is observed in their representations. We contend that this observation provides compelling evidence for the uniqueness and specialized functionality of ISNs. This finding highlights the ability of these neurons to recognize and process the core elements of instructions, with limited influence from superficial differences in expression. 

As shown in Figure \ref{fig:expert}, the activation of experts within the same type of instructions also exhibits significantly higher correlation, particularly prominent in \textit{classification}, \textit{code}, and \textit{math} tasks, which demonstrates the uniqueness of ISEs. We believe this also supports the hypothesis that different experts in a MoE model specialize in distinct skills. Through the design of the load-balancing loss, the MoE model ensures that different experts develop unique capabilities to handle instructions from different categories.

\subsection{Features of Sparse Components}

\paragraph{Similarity of Processing Instructions}

According to Figure \ref{fig:ISN_distribution}, the overall trend of the distribution of ISNs by layer remains largely unchanged before and after model fine-tuning, particularly evident in the LLaMA-2-7B, LLaMA-2-13B, and Qwen1.5-MoE-A2.7B. This observation suggests that the fundamental logic with which each model processes instructions does not undergo significant changes through fine-tuning. However, following fine-tuning, these models exhibit an increase in the number of more capable and specialized ISNs. These enhanced neurons enable the models to handle a wider variety of tasks and generate more accurate and contextually appropriate responses.

Another key similarity appears across different instruction types. Despite the substantial diversity among instructions, the distribution patterns of ISNs follow remarkably consistent trends in all tested models, especially in LLaMA-2-7B, LLaMA-2-13B, and Mistral-7B-v0.1 series. This suggests that LLMs likely rely on a shared computational mechanism for processing instructions, i.e., one where the underlying neural activation patterns remain stable regardless of instruction type.

\paragraph{Understanding ISNs Working Mechanism}
Inspired by \citet{zhao2024large}, for both types of models, we propose a three-phase mechanistic framework to elucidate ISNs operational principles.

For non-MoE models, in the early stage, the number of ISNs is significantly large, as this phase involves the encoding and processing of the shallow concepts of diverse instructions.
In the intermediate stage, instructions are further generalized and understood by the language model, leading to a sharp reduction in the number of ISNs.  
In the final stage, the number of neurons specific to particular instructions increases sharply again. These ISNs facilitate the generation of corresponding outputs by continuously decoding content into the relevant output tokens. This observation aligns with the insights proposed by \citet{huo2024mmneuron}.

For MoE models, the early stage, which extends from the shallowest layers to the intermediate layers, sees a continuous increase in the number of ISNs, enriching the representation of instructions as more experts participate in the processing. Our hypothesis is that MoE models require more steps to continuously understand and process the content of instructions. In the middle stage, the number of ISNs decreases. Similarly, in the final stage, there is another sharp increase, enabling the model to generate the corresponding outputs.

\begin{figure}[t]
    \centering 
    \includegraphics[width=0.45\textwidth]{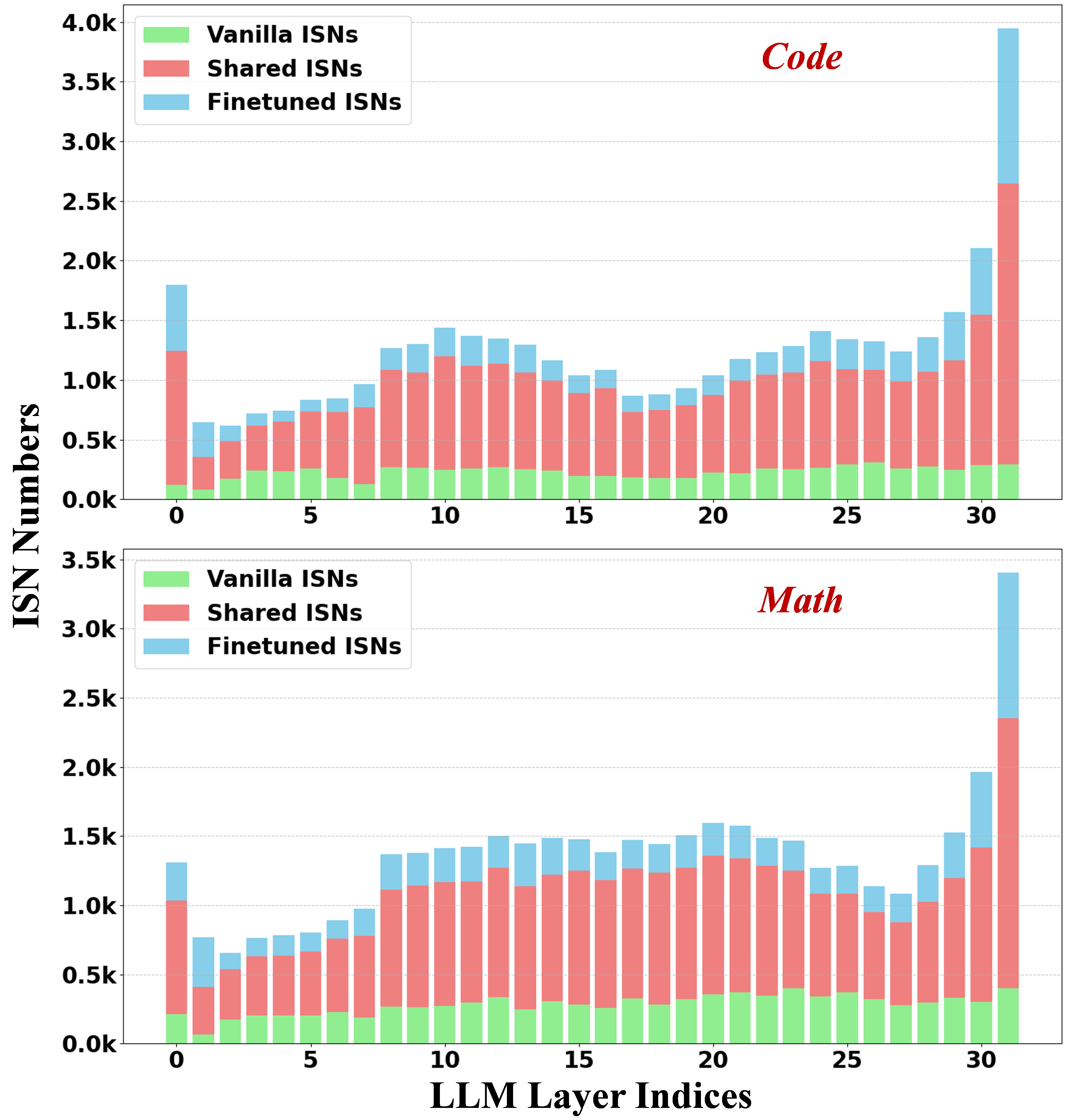}
    \caption{Venn-bar diagram illustrating the distribution changes of activated ISN numbers for two example instructions from \textit{code} and \textit{math} categories, using LLaMA-2-Chat-7B series.}
    \label{fig:venn}
    \vspace{-4mm}
\end{figure}

\subsection{Alterations in Sparse Components Following Fine-tuning}

As shown in Table \ref{tab:performance_comparison}, the activation patterns of specific neurons in response to the same instruction within the same model exhibit significant changes before and after fine-tuning. This provides additional validation for the effectiveness of our identification of ISNs. As illustrated in Figure \ref{fig:venn}, the changes in activation patterns before and after fine-tuning in LLaMA-2-7B are layer-specific: the increase in ISNs is primarily observed in the early layers (responsible for initial instruction parsing) and late layers (involved in output generation), and this pattern is consistent across two different instruction types. The overlapping neurons that inherently respond to instructions undergoes further refinement during fine-tuning. Additionally, new ISNs emerge post-fine-tuning. These newly formed and refined neurons work in tandem to establish more precise instruction-to-response mappings, demonstrating enhanced functional specialization and contributing to improved performance. This aligns closely with insights proposed by \citet{prakash2024fine}.

According to Table \ref{tab:experts_comparison}, it can be observed that before and after fine-tuning, the same instructions still activate experts with a high degree of correlation, indicating a strong linear relationship. This suggests that, from the perspective of experts, their responses remain highly consistent before and after fine-tuning. The underlying architecture and decision-making process of the model remain relatively stable, meaning that fine-tuning has not significantly altered the model's reliance on different experts. This is also reflected in \S\ref{appendix: experts top5}. Instead, the ISNs of the experts may have played a more significant role in the improved performance.

\begin{table}[t]
\centering
\resizebox{0.95\columnwidth}{!}{
\begin{tabular}{lccccccc}
\toprule
\textbf{Model} & \textbf{CLS} & \textbf{CODE} & \textbf{QA} & \textbf{GEN} & \textbf{MATH} & \textbf{SUM} \\
\midrule
LLaMA-7B Series & 0.59 & 0.62 & 0.59 & 0.56 & 0.62 & 0.57 \\
\midrule
LLaMA-13B Series & 0.55 & 0.59 & 0.56 & 0.51 & 0.59 & 0.53 \\
\midrule
Mistral-7B Series & 0.43 & 0.43 & 0.42 & 0.38 & 0.42 & 0.41 \\
\midrule
Qwen-MoE-2.7B Series & 0.44 & 0.53 & 0.59 & 0.53 & 0.51 & 0.50 \\
\bottomrule
\end{tabular}
}
\caption{Jaccard similarity coefficient in ISNs of the same instruction following fine-tuning, illustrating the Alterations in ISNs.}
\label{tab:performance_comparison}
\end{table}

\begin{table}[t]
\centering
\resizebox{0.95\columnwidth}{!}{ 
\begin{tabular}{lccccccc}
\toprule
\textbf{Model} & \textbf{CLS} & \textbf{CODE} & \textbf{QA} & \textbf{GEN} & \textbf{MATH} & \textbf{SUM} \\
\midrule
Qwen-MoE-2.7B Series & 0.91 & 0.93 & 0.94 & 0.92 & 0.91 & 0.92 \\
\bottomrule
\end{tabular}
}
\caption{Pearson correlation coefficient in ISEs of the same instruction following fine-tuning, illustrating the Alterations in ISEs.}
\label{tab:experts_comparison}
\vspace{-4mm}
\end{table}

\section{Conclusion}

This study provides novel insights into how instruction tuning shapes LLMs through sparse components. By introducing the \textbf{\dataset} dataset and \textbf{\framework} framework, we systematically identify and analyze ISNs and ISEs, revealing their unique distribution patterns and activation behaviors. Our findings advance the understanding of fine-tuning mechanisms, demonstrating how targeted modifications to sparse components significantly enhance instruction-following capabilities. This work opens new directions for interpretability research and efficient model optimization.

\clearpage
\section*{Limitations}

Our study primarily investigates the characteristics of LLMs in processing different instructions from a mechanistic explainability perspective. While we have identified certain Instruction-Specific Neurons and Experts, developing effective strategies to leverage these components to enhance the model’s instruction-following capabilities and task-solving performance is an important direction for future work. Furthermore, this paper focuses on a limited set of six representative instructions. To gain a more comprehensive understanding, future work will explore larger and more diverse datasets to identify a broader range of Instruction-Specific Neurons and Experts, thereby enhancing the generalizability of our findings.

\bibliography{acl_latex}

\clearpage
\appendix

\section{AI-Generated Instructions}
\label{appendix: AI-Generated Instructions}

\begin{tcolorbox}[
  enhanced,
  colback=blue!3!white,
  colframe=blue!40!white,
  fonttitle=\bfseries,
  title={Prompt of Instruction Generation},
  coltitle=black,
  colbacktitle=blue!40!white,
  boxrule=0.8pt,
  top=4pt,
  bottom=4pt,
  left=5pt,
  right=5pt,
]
\textbf{Classification}: You are an expert instruction generator specialized in crafting diverse SENTIMENT CLASSIFICATION tasks. Please produce exactly 100 distinct instructions. Your output must meet the following requirements: the lengths of the instructions themselves should vary from short to long and they should cover different domains.\\ 
\textbf{Code}: You are an expert instruction generator specialized in crafting diverse CODE tasks. Please produce exactly 100 distinct instructions. Your output must meet the following requirements: the lengths of the instructions themselves should vary from short to long and they should cover different levels.\\ 
\textbf{GeneralQA}: You are an expert instruction generator specialized in crafting diverse GENERALQA tasks. Please produce exactly 100 distinct instructions. Your output must meet the following requirements: the lengths of the instructions themselves should vary from short to long and they should cover different domains.\\ 
\textbf{Generation}: You are an expert instruction generator specialized in crafting diverse GENERATION tasks. Please produce exactly 100 distinct instructions. Your output must meet the following requirements: the lengths of the instructions themselves should vary from short to long and they should cover different domains. \\ 
\textbf{Math}: You are an expert instruction generator specialized in crafting diverse MATH tasks. Please produce exactly 100 distinct instructions. Your output must meet the following requirements: the lengths of the instructions themselves should vary from short to long and they should cover different levels.\\ 
\textbf{Summarization}: You are an expert instruction generator specialized in crafting diverse SUMMARIZATION tasks. Please produce exactly 100 distinct instructions. Your output must meet the following requirements: the lengths of the instructions themselves should vary from short to long and they should cover different domains.\\  
\end{tcolorbox}

\section{Dataset Post-Validation}
\label{appendix: validation}

In the post-validation process, we engaged two graduate students and one PhD student majoring in computer science. Among them, the two graduate students served as junior annotators, and the PhD student served as the senior annotator. We provided a compensation of 1000 Chinese RMB to each of the three annotators. In the initial phase of the validation process, the comprehensive review of the content was carried out by junior annotators, with a focus on error checking. This included verifying whether the instructions contained obvious grammatical or spelling errors, identifying any redundancies, detecting potential classification errors, and ensuring that the instructions were expressed clearly and unambiguously. For the aspects of classification accuracy and clarity of the instructions, we collected suggestions from both annotators. In cases where discrepancies arose, these differences were flagged and subsequently referred to our senior annotator, who was responsible for making the final adjudication. Below, we present the specific evaluation forms provided to the annotators: 

\begin{tcolorbox}[
  enhanced,
  colback=black!3!white,
  colframe=black!40!white,
  fonttitle=\bfseries,
  title={Dataset Post-Validation Process},
  coltitle=black,
  colbacktitle=black!40!white,
  boxrule=0.8pt,
  top=4pt,
  bottom=4pt,
  left=5pt,
  right=5pt,
]
\textbf{Please answer the following questions.} \\
\textbf{Instruction: } \texttt{Write a Java application to perform matrix multiplication on two-dimensional arrays.}\\
\textbf{Instruction Type: } \texttt{Code}\\
1. Whether the instructions contain obvious grammatical or spelling errors? \\ 
2. Whether the instructions contain any redundancies? \\ 
3. Whether the instructions exist potential classification errors? \\ 
4. Whether the instructions are expressed clearly and unambiguously? \\ 
\end{tcolorbox}

\section{Dataset Example}
\label{appendix: datasets}

Concrete examples are provided in Table \ref{tab:instructions_comparison} for each of the six instruction types.

\begin{table*}[ht]
\centering
\begin{tabularx}{\textwidth}{l>{\centering\arraybackslash}p{6cm}>{\centering\arraybackslash}p{6cm}}
\toprule
\textbf{Instruction Type} & \textbf{Natural Instruction} & \textbf{AI-Generated Instruction} \\
\midrule
Classification & In this task, you are given a text from tweets. Your task is to classify given tweet text into two categories: 1) positive, and 2) negative based on its content. & Can you evaluate the emotional response in this online marketplace comment? \\
\midrule
Code & From a list of integers, remove all elements that occur more than once. Keep order of elements left the same as in the input. & Write a Java application to perform matrix multiplication on two-dimensional arrays. \\
\midrule
GeneralQA & Who was the first female presenter of Blue Peter? & Who is the ancient Egyptian deity associated with the afterlife? \\
\midrule
Generation & Come up with a slogan to describe a new lipstick product. & Generate a set of ethical guidelines for the use of AI in the healthcare industry. \\
\midrule
Math & If 4 daps = 7 yaps, and 5 yaps = 3 baps, how many daps equal 42 baps? & Determine the median of the set: [22, 17, 31, 28, 19]. \\
\midrule
Summarization & Given the background description of some cooking related query, summarize the question into a title. & Generate a brief synopsis of a book on the impact of the digital revolution on traditional media and journalism. \\
\bottomrule
\end{tabularx}
\caption{Dataset examples.}
\label{tab:instructions_comparison}
\end{table*}

\begin{table*}[ht]
    \centering
    \begin{tabular}{cc}
        \toprule
        \textbf{Vanilla Models} & \textbf{Fine-tuned Models} \\
        \midrule
        LLaMA-2-7B & LLaMA-2-Chat-7B \\
        \midrule
        LLaMA-2-13B & LLaMA-2-Chat-13B\\
        \midrule
        Mistral-7B-v0.1 & Mistral-7B-Instruct-v0.1\\
        \midrule
        Qwen1.5-MoE-A2.7B & Qwen1.5-MoE-A2.7B-Chat \\
        \bottomrule
    \end{tabular}
    \caption{Vanilla and fine-tuned model names.}
    \label{tab:model names}
\end{table*}

\section{Implementation Details}
\label{appendix: implementation}

To mitigate the excessive computational time required for generation, for the Jaccard similarity calculations and Pearson correlation coefficient involved in Figures \ref{fig:jaccard} and \ref{fig:expert}, we conducted calculations on a total of 300 randomly sampled instances. Under this setup, each small square in the figures requires 2500 corresponding computations.

\section{Vanilla and Fine-tuned Models}
\label{appendix: vanilla}

We list the specific names and provide links for all vanilla and fine-tuned models in Table \ref{tab:model names} and \ref{tab:model links}.

\begin{table*}[ht]
    \centering
    \resizebox{\textwidth}{!}{
    \begin{tabular}{lc}
        \toprule
        Model & Link \\
        \midrule
        LLaMA-2-7B & \url{https://huggingface.co/meta-llama/Llama-2-7b-hf} \\
        LLaMA-2-Chat-7B & \url{https://huggingface.co/meta-llama/Llama-2-7b-chat-hf} \\
        LLaMA-2-13B & \url{https://huggingface.co/meta-llama/Llama-2-13b-hf} \\
        LLaMA-2-Chat-13B & \url{https://huggingface.co/meta-llama/Llama-2-13b-chat-hf} \\
        Mistral-7B-v0.1 & \url{https://huggingface.co/mistralai/Mistral-7B-v0.1} \\
        Mistral-7B-Instruct-v0.1 & \url{https://huggingface.co/mistralai/Mistral-7B-Instruct-v0.1} \\
        Qwen1.5-MoE-A2.7B & \url{https://huggingface.co/Qwen/Qwen1.5-MoE-A2.7B} \\
        Qwen1.5-MoE-A2.7B-Chat & \url{https://huggingface.co/Qwen/Qwen1.5-MoE-A2.7B-Chat} \\
        \bottomrule
    \end{tabular}
    }
    \caption{Model links.}
    \label{tab:model links}
\end{table*}

\section{Experts Activation Counts}
\label{appendix: experts top5}

For each instruction category, we count the total number of times each expert is activated out of two hundred instructions, and identify the top five most frequently activated experts by ID in Table \ref{tab:experts top5}. It can be observed that there is minimal variation in the top five experts across different types of instructions. Specifically, for the six types of instructions, a total of 25 experts remained in the top five.

\begin{table*}[ht]
\centering
\resizebox{\textwidth}{!}{
\begin{tabular}{llccccc|ccccc|ccccc}
\toprule
Model & Task Category & \multicolumn{5}{c}{Classification} & \multicolumn{5}{c}{Code} & \multicolumn{5}{c}{GeneralQA} \\
\midrule
\multirow{2}{*}{Qwen-MoE-2.7B-Chat} & Expert ID     & \cellcolor{myblue} 11 & \cellcolor{myblue} 52 & \cellcolor{myblue} 25 & 58 & \cellcolor{myblue} 36 & \cellcolor{myblue} 51 & \cellcolor{myblue}17 & \cellcolor{myblue} 27 & 52 & \cellcolor{myblue} 19 & \cellcolor{myblue} 52 & \cellcolor{myblue}3 & \cellcolor{myblue} 58 & 51 & \cellcolor{myblue} 42 \\
                         & Activation Counts (k)     & 12.9 & 12.8 & 12.3 & 11.9 & 11.8 & 11.8 & 11.8 & 11.5 & 11.1 & 11.1 & 6.6 & 6.0 & 6.0 & 5.8 & 5.8 \\
\midrule
\multirow{2}{*}{Qwen-MoE-2.7B} & Expert ID     & \cellcolor{myyellow} 11 & \cellcolor{myyellow} 52 & \cellcolor{myyellow} 25 & \cellcolor{myyellow} 36 & 35 & \cellcolor{myyellow} 51 & \cellcolor{myyellow}27 & \cellcolor{myyellow} 19 & \cellcolor{myyellow} 17 & 7 & \cellcolor{myyellow} 52 & \cellcolor{myyellow}42 & \cellcolor{myyellow}3 & \cellcolor{myyellow} 58 & 53 \\
                         & Activation Counts (k)     & 13.5 & 13.3 & 12.6 & 12.5 & 12.1 & 11.7 & 11.6 & 11.6 & 11.4 & 11.3 & 6.5 & 6.1 & 6.0 & 5.9 & 5.7 \\
\bottomrule
\end{tabular}
}

\medskip 

\resizebox{\textwidth}{!}{
\begin{tabular}{llccccc|ccccc|ccccc}
\toprule 
Model & Task Category & \multicolumn{5}{c}{Generation} & \multicolumn{5}{c}{Math} & \multicolumn{5}{c}{Summarization} \\
\midrule 
\multirow{2}{*}{Qwen-MoE-2.7B-Chat} & Expert ID     & \cellcolor{myblue} 52 & \cellcolor{myblue} 51 & \cellcolor{myblue}11 & \cellcolor{myblue}25 & \cellcolor{myblue}21 & 17 & 51 & \cellcolor{myblue}19 & \cellcolor{myblue}52 & \cellcolor{myblue}42 & \cellcolor{myblue}11 & \cellcolor{myblue}52 & \cellcolor{myblue}51 & \cellcolor{myblue}25 & \cellcolor{myblue}19 \\
                         & Activation Counts (k)     & 6.0 & 6.0 & 5.6 & 5.2 & 5.1 & 11.7 & 11.6 & 11.4 & 11.4 & 11.3 & 11.5 & 10.5 & 10.1 & 10.1 & 9.3 \\
\midrule
\multirow{2}{*}{Qwen-MoE-2.7B} & Expert ID     & \cellcolor{myyellow}52 & \cellcolor{myyellow}51 & \cellcolor{myyellow}11 & \cellcolor{myyellow}25 & \cellcolor{myyellow}21 & \cellcolor{myyellow}42 & \cellcolor{myyellow}52 & \cellcolor{myyellow}19 & 7 & 22 & \cellcolor{myyellow}11 & \cellcolor{myyellow}52 & \cellcolor{myyellow}25 & \cellcolor{myyellow}19 & \cellcolor{myyellow}51 \\
                         & Activation Counts (k)     & 6.1 & 5.9 & 5.6 & 5.4 & 5.3 & 11.9 & 11.7 & 11.7 & 11.5 & 11.5 & 12.0 & 10.9 & 10.5 & 9.9 & 9.8 \\
\bottomrule
\end{tabular}
}

\caption{Top five experts activation counts by instruction types. The highlighted experts represent those that remain among the top five most frequently activated experts before and after fine-tuning.}
\label{tab:experts top5}

\end{table*}

\end{document}